\pdfoutput=1

\documentclass[11pt,a4paper]{article}
\usepackage[hyperref]{acl2021}
\usepackage{times}
\usepackage{latexsym}
\usepackage{float}
\usepackage{amsmath}
\usepackage{graphicx}

\usepackage{microtype}

\aclfinalcopy 

\title{Gender Bias Hidden Behind Chinese Word Embeddings:\\The Case of Chinese Adjectives}

\author{Meichun Jiao, Ziyang Luo\\
  {\normalsize Department of Linguistics and Philology, Uppsala University, Sweden} \\
  \texttt{\normalsize \{Meichun.Jiao.1608,Ziyang.Luo.9588\}@student.uu.se}}

\date{}

\begin{document}
\maketitle
\begin{abstract}
Gender bias in word embeddings gradually becomes a vivid research field in recent years. Most studies in this field aim at measurement and debiasing methods with English as the target language. This paper investigates gender bias in static word embeddings from a unique perspective, Chinese adjectives. By training word representations with different models, the gender bias behind the vectors of adjectives is assessed. Through a comparison between the produced results and a human scored data set, we demonstrate how gender bias encoded in word embeddings differentiates from people’s attitudes.
\end{abstract}

\noindent\textbf{BIAS STATEMENT}\indent This paper studies gender bias in Chinese adjectives, captured by word embeddings. For each Chinese adjective, a gender bias score is calculated by $\vec{w}\cdot(\vec{he}-\vec{she})$ \citep{bolukbasi2016man}. A positive score represents the Chinese adjective word embeddings is more associated with males, and a negative value refers to the opposite result. In our daily life, we find that gender stereotypes can be conveyed by adjectives. The close association between an adjective and a certain gender could be the accomplice in forming gender stereotypes \citep{menegatti2017gender}. If these stereotypes are learned by the adjective word embeddings, they would be propagated to downstream NLP applications; accordingly, the gender stereotypes would be reinforced in users' mind. For example, the system will tend to use ``smart'' to describe males because of the existed social stereotype in training data that males are good at mathematics; then, the influence of the stereotype would be spread and increased again. Thus, we want to further investigate the bias encoded by the embeddings and how they are different with what in people's mind. 

\section{Introduction}

In the deep learning era, a major area of NLP has concerned the representation of words in low-dimensional and continuous vector spaces. People propose different algorithms to achieve such goal, including Word2Vec \citep{mikolov2013efficient}, GloVe \citep{pennington-etal-2014-glove} and FastText  \citep{bojanowski-etal-2017-enriching}. Word embeddings play an important role in many NLP tasks, such as machine translation \citep{qi-etal-2018-pre} and sentiment analysis \citep{yu-etal-2017-refining}. However, several studies point out that word embeddings could capture the gender stereotypes in training data and transmit them to downstream applications \citep{bolukbasi2016man,zhao2017men}. The consequence is often unbearable. Take machine translation as an example, if we translate a sentence concerning ``nurse'' from a language with gender-neutral pronouns to English, a female pronoun might be automatically produced to denote ``nurse'' \citep{prates2019assessing}. Undoubtedly, this falls into the trap of the typical gender stereotypes. Therefore, the investigation of gender bias in word embeddings is necessary and accordingly attracts scholars’ attention in recent years \citep{bolukbasi2016man,zhao2017men}.

Most previous studies concerning gender bias in word embeddings only take English as the target language. Other languages are only included in several multi-lingual projects. For example, \citet{prates2019assessing} evaluate the gender bias in machine translation by translating 12 gender neutral languages with the Google Translate API; \citet{lewis2020gender} examine whether gender stereotypes could be reflected in the large-scale distributional structure of 25 natural languages. Apart from English, other languages have rarely been the target language in the research under this topic. This paper will take Chinese as the target language, investigating gender bias in word embeddings trained with the model designed for special features of Chinese.

The fact that social stereotypes are conveyed in our language is often neglected by the public. From the commonly used adjectives, we could get a glimpse of the social stereotypes of a certain group of people. These stereotypes would confine us to what we should be in the minds of the public. It has been confirmed that when describing different genders, people will choose divergent groups of adjectives even though such a choice might change with the development of society \citep{2018Word}. Therefore, this study focuses on the problem of gender stereotypes from the perspective of adjectives. By scoring the gender bias from our trained vectors, we yield a subjective result of the gender preference of a set of adjectives. Through comparing our results with a handcrafted data set of human attitudes towards adjectives\citep{zhu2020}, we find that what is encoded in word embeddings is, to some extent, inconsistent with people’s feelings on the gender preference of these adjectives. 

\section{Related work}

Gender could affect the usage of adjectives \citep{lakoff1973language}. On the other hand, the attitude of the public towards the social roles of men and women could also be indicated by how adjectives correlates with genders\citep{zhu2020}. In the past decade, an increasing number of studies investigating adjectives and gender stereotypes from various perspectives are proposed and developed. \citet{baker2013will} reveals the stereotype in the description of boys and girls by analyzing adjectives only used for a certain gender with the aid of corpora covering a range of written genres. Research of Bollywood movies \citep{madaan2018analyze} finds that different adjectives are chosen when they try to create impressive male and female roles. The significant divergence between the usage of adjectives for describing men and women has also been confirmed by \citet{hoyle2019unsupervised}, and they also notice the variance is consistent with common stereotypes. \citet{zhu2020} trace the change of gender bias in Chinese adjectives based on a handcrafted data set that consists of the gender preference score of adjectives. However, the number of studies focusing on Chinese adjectives and gender bias is still limited.

Gender bias in word embeddings and the corresponding debiasing methods have been a vivid research field in recent years. \citet{bolukbasi2016man} and \citet{caliskan2017semantics} confirm that word embedding models could precisely capture the social stereotypes concerning people’s careers, such as the relationship in an analogy that \textit{Man is to Computer Programmer as Woman is to Homemaker}. This bias could even be amplified by embedding models \citep{zhao2017men}. Besides English, other target languages like Swedish \citep{sahlgren-olsson-2019-gender} and Dutch \citep{wevers2019using} gradually attract the attention of researchers. Various methods for assessing bias and debiasing are proposed and developed in previous studies. \citet{bolukbasi2016man} firstly measure the gender bias by computing the projection of a word on $\vec{he} - \vec{she}$ direction, which has been confirmed strongly correlated with the public judgment of gender stereotypes. Based on this method, they also develop a debiasing method by post-processing the generated word vectors. \citet{zhao2018learning} and \citet{zhang2018mitigating} further propose to debias word embeddings in training procedure by changing the loss of GloVe model \citep{pennington2014glove} and employing an adversarial network, respectively. Despite a large amount of research having been done in this field, to the best of our knowledge, no one has assessed the underlying gender bias behind adjectives, especially those in non-English languages. 

To complement the full picture of gender bias encoded in word representation, this paper examines the problem from the perspective of adjectives rather than nouns of occupations that repeatedly appeared in previous studies. Based on the human scoring data set of \citet{zhu2020}, we investigate the similarities and differences between the automatically captured gender bias in Chinese and people’s judgement. 

\section{Methodology}
To uncover the gender stereotypes conveyed by adjectives, we first preprocess a corpus of online Chinese news and train word embeddings on it with two different models. Then, we calculate the gender bias scores based on the generated two vectors and compare them with the human scoring data set, Adjectives list with Gendered Skewness and Sentiment (AGSS) \citep{zhu2020}.

\subsection{Data}

News reports are not only the reflection of social consciousness but also the easily collected corpus for many NLP tasks. Therefore, we choose a corpus of Chinese news reports as our training data set. It was collected and released by Sogou Labs, covering 18 themes of news from various Chinese news websites.\footnote{http://www.sogou.com/labs/resource/ca.php} The details of the corpus are illustrated in Table \ref{tab:corpus-detail}. All texts in the data set are cleaned and preprocessed through the following steps.
\begin{enumerate}
    \item Extract the news content and change the encoding from gbk to utf-8. All the other information and metadata are removed. 
    \item Remove the html tag by the regular expression and conduct Chinese word segmentation with \textit{jieba},\footnote{https://github.com/fxsjy/jieba} a widely used Python module.
\end{enumerate}

\begin{table}[]
    \centering
    \begin{tabular}{l|l}
        \hline
        Original size & 1.54GB\\
        Size after preprocessing & 2.1GB\\
        The number of tokens & 375.3M\\
        The number of unique words & 100.7k\\
        \hline
    \end{tabular}
    \caption{The details of the Chinese news corpus.}
    \label{tab:corpus-detail}
\end{table}

\subsection{Training and evaluation of word embeddings}

\begin{figure}
    \centering
    \includegraphics[width=4cm]{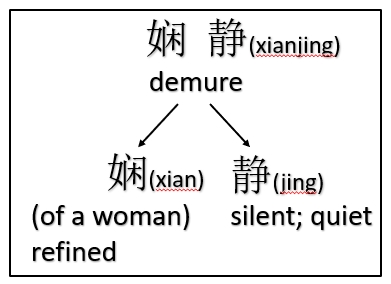}
    \caption{An example of semantic relation between Chinese words and characters. Pinying (pronunciation of the word or character) is in the lower right parentheses; English translation is noted directly below the word or character}
    \label{xianjing}
\end{figure}

The meaning of Chinese words is usually related to the semantic information carried by the characters (Hanzi) that they are comprised of. Figure \ref{xianjing} shows an example: the word ``xianjing'' means ``demure'', which consists of two characters. The first one, ``xian'', means refined but usually used for describing a woman; the second character ``jing'' means silent and quiet. The word inherits and combines the meaning of each character, even the information concerning gender. This feature of Chinese leads to the development of word embedding models in which word vectors are trained with the character-level information. However, no study before has provided any ideas about how the encoding of gender bias information will be affected by training embedding with character-level information. Therefore, we decide to train our vectors with one of such models, namely the character-enhanced word embedding model (CWE) \citep{chen2015joint}. In addition to the word vector, this model also trains a vector for Chinese characters.

CWE is developed based on the framework CBOW \citep{mikolov2013distributed}. CBOW aims at predicting the target word by understanding the surrounding context words. Practically, its objective is to maximize the average log probability given a word sequence $D=\left\{x_{1}, \ldots, x_{M}\right\}$. CWE modifies the way of representing the context words in the algorithm of CBOW, predicting target words by combining character embedding and word embedding. A context word $\mathbf{x}_{j}$ in CWE would be represented as follows,
\begin{equation}
    \mathbf{x}_{j}=\frac{1}{2}\left(\mathbf{w}_{j}+\frac{1}{N_{j}} \sum_{k=1}^{N_{j}} \mathbf{c}_{k}\right).
\end{equation}
$\mathbf{w}_{j}$ refers to the word embedding of $\mathbf{x}_{j}$; $N_{j}$ represents the number of characters in $\mathbf{x}_{j}$; $\mathbf{c}_{k}$ is the representation of the $\mathbf{k}$-th character in $\mathbf{x}_{j}$. For comparison, we also train vectors on CBOW to show in the influence of character-level information. The Python library \textit{Gensim} \footnote{https://github.com/RaRe-Technologies/gensim} is used for training the representation with CBOW, and the other with CWE is completed by the released code of \citet{chen2015joint}.\footnote{https://github.com/Leonard-Xu/CWE/tree/master/src} To make the results comparable, we keep the same hyper-parameters for the two models. Detailed information is recorded in Table \ref{tab:training_detail}.

\begin{table}
    \centering
    \begin{tabular}{l|l}
        \hline
        Window size & 5\\
        Iteration & 5\\
        Dimension & 300\\
        Min\_count & 8\\
        Num\_threads & 12\\
        \hline
    \end{tabular}
    \caption{Word embeddings training hyper-parameter details.}
    \label{tab:training_detail}
\end{table}

To ensure the effective of the produced embeddings, we evaluate them by word analogy tasks and the corresponding tools developed by \citet{li-etal-2018-analogical}. The test data set of the task includes 17813 questions about morphological or semantic relations. \footnote{https://github.com/Embedding/Chinese-Word-Vectors/tree/master/testsets} The results are illustrated in table \ref{tab:eva-results}. Although the semantic task results are lower than the values given in the paper of \citet{li-etal-2018-analogical}, we still assume that they are reliable as the size of our training data is only the half of theirs.

\begin{table}[htp]
    \centering
    \begin{tabular}{lcc}
        \hline
        Model & Morphological & Semantic\\
        \hline
        \hline
        \citet{li-etal-2018-analogical} & 11.5 & 30.2 \\
        \hline
        \hline
        CBOW  & 11.1 & 23.5 \\
        CWE & 19.7 & 24.6 \\
        \hline
\end{tabular}
\caption{Accuracy scores of different word embeddings in the evaluation tasks. The results are reported as $acc\times 100$.}
\label{tab:eva-results}
\end{table}

\subsection{Gender bias measurement and data set}\label{sec:3.3}
We employ the method of \citet{bolukbasi2016man} to assess gender bias encoded in the trained embeddings. For each adjective, a gender bias score is calculated by $\vec{w}\cdot(\vec{he} - \vec{she})$ based on its vector.\footnote{We use the Chinese translation of he and she when conducting experiments.} A positive result presents that the word has a closer association with males, while a negative score implies that the word is more associated with females. The higher the absolute value, the more biased the adjective is. 0 means totally neutral. 

Adjectives List with Gendered Skewness and Sentiment (AGSS) is a handcrafted data set built by questionnaire in the project of \citet{zhu2020}. 
6 linguists firstly select 466 Chinese adjectives that could describe people, then 116 gender-balanced respondents score these adjectives by questionnaires. The the scale of score 1 to 5 is used to reflect people’s attitude, with 1 being more related to female and 5 more related to male. 
Table \ref{AGSS} shows some example data from AGSS. Finally, 304 adjectives are scored larger than 3, 153 adjectives get score less than 3, and 9 are believed totally neutral. According to the statistics of AGSS, the adjectives  chosen for this data set are more associated with males, so \citet{zhu2020} state that AGSS is with gender skewness. To analyze the results, we compare our gender bias scores from word embeddings with the AGSS scores. 
As they are on different scales, Pearson correlation coefficient is employed here. It could measure the the strength of the linear association between two variables, which returns a value between -1 and 1. $1$ indicates strong positive linear correlation, 0 indicates no linear correlation and $-1$ indicates a strong negative linear correlation.

\begin{table}
    \center
    \begin{tabular}{lc}
        \hline
        Words & Gender skewness in AGSS\\
        \hline
        \hline
        powerful & 4.44\\
        vuglar & 3.62\\
        selfless & 3.00\\
        cute & 2.26\\
        decorous & 1.59\\
        \hline
    \end{tabular}
    \caption{Example data from AGSS. Each word is translated into English.}
    \label{AGSS}
\end{table}

\section{Results and discussion}
\subsection{Gender bias scores from word embeddings}
We calculate the gender bias score for the same adjectives with AGSS and conclude the basic statistics in Table \ref{tab:gb-results}. More adjectives are categorized into the group close to male. This is identified with what \citet{zhu2020} state about AGSS (mentioned in Section \ref{sec:3.3}). However, it should be noticed that the average scores of both models result in a negative value. This might suggest that most absolute values of negative gender bias scores are much higher than the positive group.

\begin{table}[htp]
    \centering
    \begin{tabular}{lcc}
        \hline
        & CBOW & CWE \\
        \hline
        \hline
        \# pos. score & 283 & 316 \\
        \# neg. score & 183 & 150 \\
        Avg. score & -0.02029 & -0.02945 \\
        \hline
    \end{tabular}
    \caption{Statistics of the gender bias scores from two embeddings.}
    \label{tab:gb-results}
\end{table}

\subsection{Correlation between word vectors and AGSS}

\begin{figure*}[h]
\centering
\includegraphics[width=12cm]{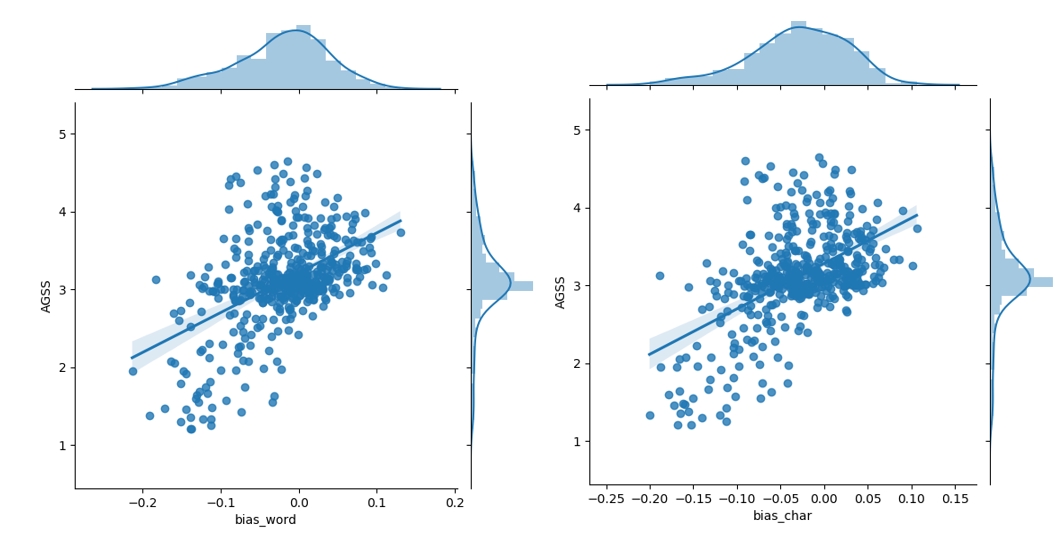}
\caption{Scatter plots of AGSS scores and gender bias scores from word vectors trained with CBOW (left) and CWE (right). AGSS refers to the AGSS scores and bias\_word and bias\_char refers to the generated gender bias scores. The distribution of gender bias scores and AGSS scores are on the top and right of the plots respectively. The lines show the linear relation between the two categories.}
\label{f1}
\end{figure*}

The Pearson correlation coefficients presented in Table \ref{tab:pearson1} suggest the two categories of data are positively associated. However, the correlation is not that strong with only around 0.5, since the range of Pearson coefficient is from -1 to 1. Besides, the gender bias scores from the word embeddings trained with CWE are more associated with the human scores. This might suggest that the character-level information could help the model capture gender bias more precisely, or we should say such information could contribute to encoding what is in people’s minds.

\begin{figure*}[htp]
\centering
\includegraphics[width=12cm]{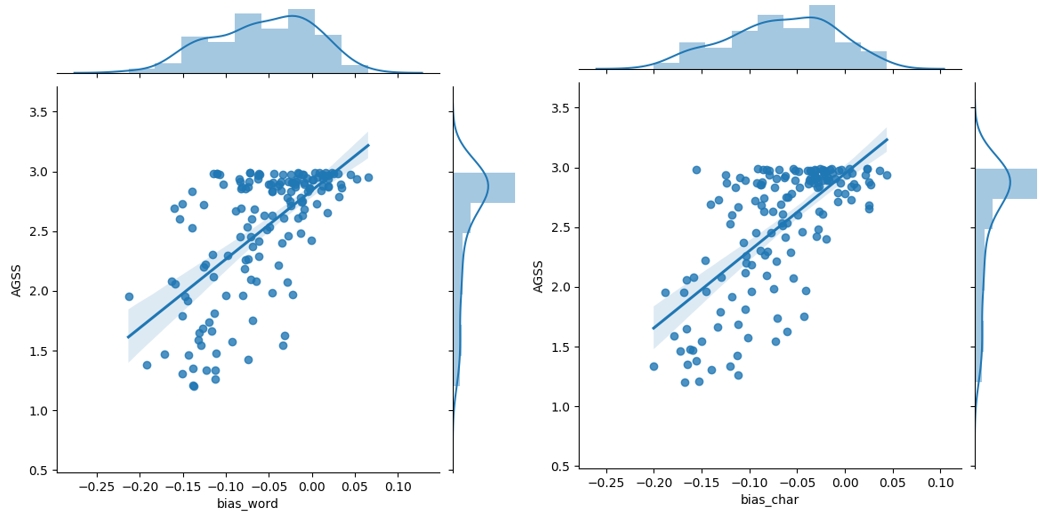}
\caption{Scatter plots of the data group with AGSS scores \textless 3. AGSS refers to the AGSS scores and bias\_word and bias\_char refers to the generated gender bias scores.}
\label{f2}
\end{figure*}

\begin{figure*}[htp]
\centering
\includegraphics[width=12cm]{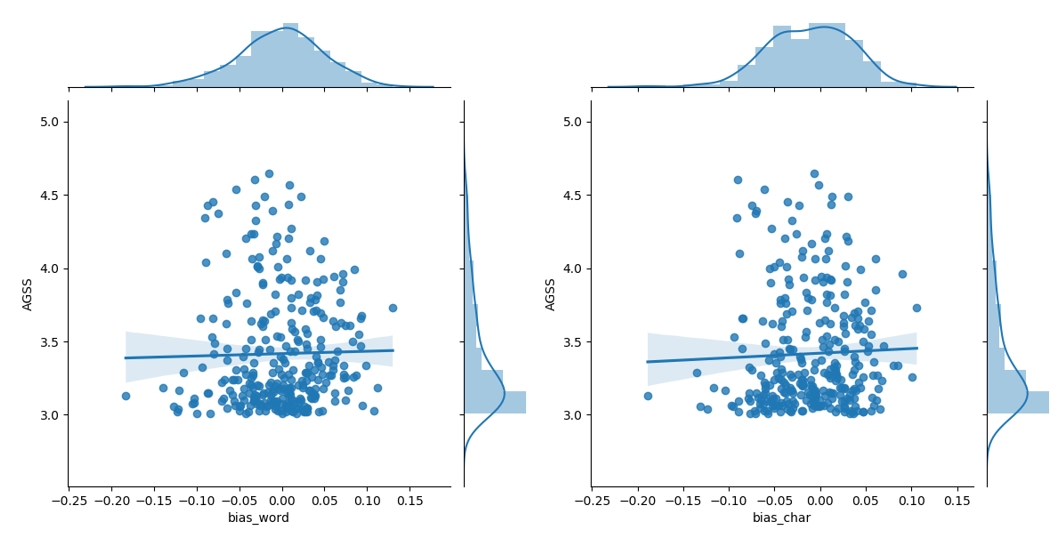}
\caption{Scatter plots of the data group with AGSS scores \textgreater 3. AGSS refers to the AGSS scores and bias\_word and bias\_char refers to the generated gender bias scores.}
\label{f3}
\end{figure*}

\begin{table}[]
    \centering
    \begin{tabular}{ccc}
        \hline
        & CBOW & CWE\\
        \hline
        \hline
        \begin{tabular}[c]{@{}c@{}}Pearson \\ coefficient\end{tabular} & 0.489 & 0.503 \\
        p-value & 0.000 & 0.000 \\
        \hline
    \end{tabular}
    \caption{Pearson correlation coefficient between AGSS score and gender bias scores from trained vectors. CBOW score and CWE score refer to the gender bias score from word vectors trained with CBOW and CWE model.}
    \label{tab:pearson1}
\end{table}

In Figure \ref{f1}, we can find more details of the correlation between the two categories of data. By comparing the distribution of the two types of scores, we notice that the scores given by people are very concentrated between 2.5 to 3.5, while automatically calculated scores have a wider distribution. This might be caused by different scales, but may also come from people hypocrisy: they spontaneously narrow the extent of gender preference of words when they are asked to score their attitudes. Besides, it is a clear tendency that some words only for males in people’s impression are automatically given a negative score, which means they are more close to women in word vectors. Therefore, we conduct further analysis by separating the data into two groups based on the neutral line in AGSS. 
\begin{table}[htp]
    \centering
    \begin{tabular}{ccc}
        \hline
        & CBOW & CWE\\
        \hline
        \hline
        \begin{tabular}[c]{@{}c@{}}Pearson \\ coefficient\end{tabular} & 0.673 & 0.628 \\
        p-value & 0.000 & 0.000 \\
        \hline
    \end{tabular}
    \caption{Pearson correlation coefficient of the data group with AGSS scores \textless 3. }
    \label{tab:pearson2}
\end{table}

\begin{table}[htp]
    \centering
    \begin{tabular}{ccc}
        \hline
        & CBOW & CWE\\
        \hline
        \hline
        \begin{tabular}[c]{@{}c@{}}Pearson \\ coefficient\end{tabular} & 0.036 & 0.020 \\
        p-value & 0.543 & 0.724 \\
        \hline
    \end{tabular}
    \caption{Pearson correlation coefficient of the data group with AGSS scores \textgreater 3. }
    \label{tab:pearson3}
\end{table}

We recalculate the Pearson correlation coefficients for the two group of data, presenting results in Table \ref{tab:pearson2} and Table \ref{tab:pearson3}. To give a full picture, separated scatter plots as shown in Figure \ref{f2} and Figure \ref{f3} are also included. The increment of coefficients for the group with AGSS scores lower than 3 suggests that most adjectives believed for describing women are closer to females in word vectors as well. What is encoded by word embedding is consistent with people's impressions of these words. In addition, the correlation for scores from vectors trained with CBOW exceeds the results with the CWE model. This finding might indicate the underlying negative influence of covering character-level information in the word embedding. 

However, a substantial divergence appears in the other group. Based on the scatter plot and the Pearson coefficient, some of the adjectives that almost exclusively connect with male in people’s minds could be very neutral according to our word embedding. The coefficients also suggest that the two categories of data do not show linear relations. Additionally, only one-third of the adjectives in this group are closer to males in word embedding, while the others are actually more associated with females. Obviously, what we estimate from embedding disagrees with people’s attitudes. This could be explained by the development of language. The study of \citet{zhu2020} proves that some Chinese adjectives for describing men in past time gradually become neutral in written language. Since the language used online develops fast and our training data are online news reports, the word embedding we trained is likely affected by the change. However, the public has not realized such development although they might start to use it in the new way. Therefore, when they are queried about the attitude towards attitudes, they might give an answer based on their outdated knowledge. 

\section{Conclusion}
In this paper, we investigate gender bias in Chinese word embeddings from the perspective of adjectives, and compare automatically calculated gender bias score with human attitudes. We elaborately present the differences between gender bias encoded in word vectors and the people’s feeling of the same adjective. For the words that people believe for describing women, the extracted score of gender bias gives an identified results; while for adjectives that should be used for men in people's mind, our results suggest that these group of words are actually more neutral than the crowd judgement. Additionally, how the word embedding models covering character-level information perform in terms of capturing gender bias in Chinese is also examined.\\

\paragraph{Acknowledgments}
This project grew out of a master course project for the Fall 2020 Uppsala University 5LN714, \textit{Language Technology: Research and Development}. We would like to thank Sara Stymne and Ali Basirat for some great suggestions and the anonymous reviewers for their excellent feedback.



\bibliographystyle{acl_natbib}
\bibliography{anthology,acl2021}


\end{document}